\documentclass{llncs}
\usepackage{llncsdoc}
\usepackage{makeidx}  % allows for indexgeneration
\usepackage{url}
\usepackage[dvips]{graphicx}
\usepackage{sidecap}
\usepackage{mdframed}
\usepackage{algorithm,algorithmic}
\usepackage{verbatim}
\usepackage{enumitem}
\usepackage[authoryear,round]{natbib}
\usepackage{hyperref}
\renewcommand\bibsection%
{
  \section*{Bibliography}
}
\usepackage{amsmath,amssymb,amsfonts,bm} 	% Math stuff
\usepackage[latin1]{inputenc}

\newcommand{\qu}[1]{``#1''}

\newcommand{\Set}[1]{\mathcal{#1}}
\def\N{\mathrm{I\kern-0.4ex N}}
\def\R{\mathrm{I\kern-0.4ex R}}
\def\E{\mathrm{I\kern-0.4ex E}}

\newcommand{\Lt}[1]{L^2(#1)}
\newcommand{\Lx}[0]{\Lt{\Set X,\indidist}}

\newcommand{\indiroot}[0]{\vartheta}

\newcommand{\indipol}[0]{\tau}
\newcommand{\indifun}[0]{\phi}

\newcommand{\basefun}[0]{\psi}
\newcommand{\indi}[0]{\indiroot}

\newcommand{\mat}[1]{\mathbf{#1}}
\newcommand{\ve}[1]{\bm #1} %{\bm{#1}}
\newcommand{\shallbe}[0]{\stackrel{\scriptscriptstyle!}{=}}

\newcommand{\smallsum}[2]{ {\textstyle 
	\sum\limits_{\scriptscriptstyle #1}^{\scriptscriptstyle #2}} 
}
\newcommand{\smallprod}[2]{ {\textstyle 
	\prod\limits_{\scriptscriptstyle #1}^{\scriptscriptstyle #2}} 
}
\newcommand{\smallfrac}[2]{ {\textstyle \frac{#1}{#2}} }

\newcommand{\wsp}[3]{\langle #1 , #2 \rangle_{ #3 }}

\newcommand{\op}[1]{{\rm \hat #1}}

\newcommand{\meanpolicy}[0]{\op \Gamma{\kern-0.5ex_\pi}}
\newcommand{\meanaction}[0]{\op \Gamma{\kern-0.5ex_\indipol}}

\newcommand{\grad}[0]{\nabla}
\newcommand{\subgrad}[1]{\grad\kern-0.5ex_{#1}}

\newcommand{\kernel}[0]{\kappa}
\newcommand{\hilbert}[0]{\Set H}

\newtheorem{Corollary}[theorem]{Corollary}
\newtheorem{Proposition}[theorem]{Proposition}

\newcommand{\cost}[0]{\Set C}
\newcommand{\acost}{{\tilde\cost}}
\newcommand{\LFF}[1]{\Set F^{#1}}
\newcommand{\FF}{\Set F}
\newcommand{\LFa}{\Set L_{\indifun}^{k}}
\renewcommand{\Lx}[0]{\Lt{\Set X, \indi}}
\newcommand{\rnd}[0]{{\scriptstyle \frac{\indi}{\xi}}}
\newcommand{\ddb}[1]{{\textstyle \frac{\partial}{\partial \ve b^{#1}}}}
\newcommand{\iteration}{{\hat{\imath}}}
\begin{document}
\frontmatter          % for the preliminaries
\pagestyle{headings}  % switches on printing of running heads

\mainmatter              % start of the contributions
\title{Regression with Linear Factored Functions}
\titlerunning{Regression with LFF}  % abbreviated title (for running head)
\author{Wendelin B\"ohmer \and Klaus Obermayer}
\authorrunning{B\"ohmer and Obermayer} % abbreviated author list (for running head)
\tocauthor{Wendelin B\"ohmer and Klaus Obermayer}
\institute{Neural Information Processing Group,
	Technische Universit\"at Berlin, \\
	Sekr.~MAR5-6, Marchstr. 23, D-10587 Berlin, Germany \\
	\email{\{wendelin,oby\}@ni.tu-berlin.de}, \quad
	\texttt{http://www.ni.tu-berlin.de}
}

\maketitle              % typeset the title of the contribution

\begin{abstract}
Many applications that use empirically estimated functions
face a {\em curse of dimensionality}, 
because the integrals over most function classes 
must be approximated by sampling.
This paper introduces a novel {\em regression}-algorithm
that learns {\em linear factored functions} (LFF).
This class of functions has structural properties 
that allow to analytically solve certain integrals 
and to calculate point-wise products.
Applications like {\em belief propagation} and 
{\em reinforcement learning} can exploit 
these properties to break the curse and speed up computation.
We derive a regularized greedy optimization scheme, 
that learns factored basis functions during training.
The novel regression algorithm performs competitively
to {\em Gaussian processes} on benchmark tasks,
and the learned LFF functions are 
with 4-9 factored basis functions on average very compact.
\end{abstract}

% ==============================================================================
\section{Introduction}
This paper introduces a novel regression-algorithm,
%based on {\em factored functions}. It 
which performs competitive to {\em Gaussian processes},
but yields {\em linear factored functions} (LFF).
These have outstanding properties
like analytical {\em point-wise products} and {\em marginalization}.

Regression is a well known problem, 
which can be solved by many non-linear architectures
like {\em kernel methods} \citep{Taylor04}
or {\em neural networks} \citep{Haykin98}.
While these perform well, 
the estimated functions often suffer
a {\em curse of dimensionality} in later applications.
For example, 
computing an integral over a neural network 
or kernel function requires to sample the entire input space.
Applications like {\em belief propagation} \citep{Pearl88} and
{\em reinforcement learning} \citep{Kaelbling96},
on the other hand, 
face large input spaces and require therefore efficient computations.
We propose LFF for this purpose
and showcase its properties in comparison to kernel functions.

% ------------------------------------------------------------------------------
\subsection{Kernel regression}

In the last 20 years, 
kernel methods like {\em support vector machines} 
\citep[SVM,][]{Vapnik95,Boser92}
have become a de facto standard in various practical applications.
This is mainly due to a sparse representation of the
learned classifiers with so called {\em support vectors} (SV).
The most popular kernel method for regression, 
{\em Gaussian processes} \citep[GP, see][]{Rasmussen06,Bishop06},
on the other hand, requires as many SV as training samples. 
Sparse versions of GP aim thus for a small subset of SV.
Some select this set based on 
constraints similar to SVM \citep{Vapnik95,Tipping01},
while others try to conserve the spanned linear function space
\citep[{\em sparse GP},][]{Csato02, Rasmussen06}.
There exist also attempts to construct 
new SV by averaging similar training samples
%\citep[e.g.~in {\em Budget SVM},][]{Wang12}.
\citep[e.g.][]{Wang12}.  

Well chosen SV for regression are usually not sparsely
concentrated on a decision boundary as they are for SVM.
In fact, many practical applications report 
that they are distributed uniformly in the input space
\citep[e.g.~in][]{Boehmer13}.
Regression tasks restricted to a 
small region of the input space may tolerate this,
but some applications require predictions everywhere.
For example, 
the {\em value function} in reinforcement learning
must be generalized to each state.
The number of SV required to {\em represent} this function
equally well in each state grows
exponentially in the number of input-space dimensions,
leading to Bellman's famous curse of dimensionality \citep{Bellman57}.

Kernel methods derive their effectiveness from 
linear optimization in a non-linear 
{\em Hilbert space} of functions.
Kernel-functions parameterized by SV are
the non-linear {\em basis functions} in this space.
Due to the functional form of the kernel,
this can be a very ineffective way to select basis functions, though.
Even in relatively small input spaces, 
it often takes hundreds or thousands SV
to approximate a function sufficiently.
To alleviate the problem, 
one can construct complex kernels out of simple prototypes
%\citep[see a recent review in][]{Goenen11}.
\citep[see e.g.][]{Goenen11}.

% ------------------------------------------------------------------------------
\subsection{Factored basis functions}

Diverging from all above arguments,
this article proposes a more radical approach: 
{to construct the non-linear {basis functions} 
directly during training, without the detour 
over kernel functions and support vectors}.
This poses two main challenges: 
to select a {\em suitable functions space}
and to {\em regularize the optimization} properly.
The former is critical, 
as a small set of basis functions must
be able to approximate any target function,
but should also be easy to compute in practice.

We propose {\em factored functions} $\basefun_i \in \FF$
as basis functions for regression,
and call the linear combination of $m$ of those bases
a {\em linear factored function} $f \in \LFF{m}$
(LFF, Section \ref{sec:lff}).
Due to their structure, 
LFF can solve certain integrals analytically
and allow very efficient computation of 
point-wise products and marginalization.
We show that LFF are universal function approximators 
and derive an appropriate
{\em regularization} term.
This regularization promotes smoothness, 
but also retains a high degree of variability in
densely sampled regions by linking smoothness to uncertainty 
about the sampling distribution.
Finally, we derive a novel regression algorithm for LFF
based on a greedy optimization scheme.

Functions learned by this algorithm 
are very compact (between $3$ and $12$ bases on standard benchmarks)
and perform competitive with Gaussian processes (Section \ref{sec:evaluation}).
The paper finishes with a discussion of 
the computational possibilities of LFF 
in potential areas of application and
possible extensions to {\em sparse regression} with LFF
(Section \ref{sec:discussion}).

% ==============================================================================
\section{Regression}

Let $\{\ve x_t \in \Set X \}_{t=1}^n$ 
be a set of $n$ {\em input samples}, 
i.i.d.~drawn from an input set $\Set X \subset \R^d$. 
Each so called \qu{training sample} is {\em labeled}
with a real number $\{y_t \in \R\}_{t=1}^n$.
{\em Regression} aims to find a function $f: \Set X \to \R$,
that predicts the labels to all
(previously unseen) test samples as good as possible.
Labels may be afflicted by {\em noise} and 
$f$ must thus approximate the mean label of each sample,
i.e., the function $\mu: \Set X \to \R$.
It is important to notice that {\em conceptually} 
the noise is introduced by two (non observable) sources: 
noisy labels $y_t$ and noisy samples $\ve x_t$.
The latter will play an important role for regularization.
We define the conditional distribution $\chi$ 
of observable samples $\ve x \in \Set X$ 
given the non-observable \qu{true} samples $\ve z \in \Set X$,
which are drawn by a distribution $\xi$.
In the limit of infinite samples,
the {\em least squares} cost-function $\cost[f|\chi,\mu]$
can thus be written as
\begin{equation} \label{eq:noisy_regression}
\lim_{n \to \infty}
\inf_{f} \frac{1}{n} \sum_{t=1}^n 
	\Big(f(\ve x_t) - y_t\Big)^2
\quad = \quad
\inf_{f}
	\iint \xi(d\ve z) \, \chi(d\ve x|\ve z)
	\Big( f(\ve x) - \mu(\ve z) \Big)^2 \,.
\end{equation}
The cost function $\cost$ can never be computed {\em exactly},
but {\em approximated} using the training samples\footnote{
	The unknown distribution $\xi$
	will be approximated with the sampling distribution.
} and assumptions about the unknown noise distribution $\chi$.

% ==============================================================================
\section{Linear factored functions} \label{sec:lff}
Any non-linear function can be expressed as a {linear function}
$f(\ve x) = \ve a^\top \ve \basefun(\ve x)$, 
$\forall \ve x \in \Set X,$
with $m$ non-linear {basis functions} 
$\basefun_i: \Set X \to \R, \, \forall i \in \{1\,\ldots,m\}$.
In this section we will define {\em linear factored functions} (LFF),
that have {\em factored basis functions} 
$\basefun_i(\ve x) := \basefun^1_i(x_1) \cdot 
\ldots \cdot \basefun_i^d(x_d) \in \FF$,
a regularization method for this function class 
and an algorithm for regression with LFF.

% ------------------------------------------------------------------------------
\subsection{Function class} \label{sec:lff_definition}
We define the class of { linear factored functions} $f \in \LFF{m}$
as a linear combination (with linear parameters $\ve a \in \R^m$) of 
$m$ factored basis functions $\basefun_i: \Set X \to \R$ 
(with parameters $\{ \mat B^k \in \R^{m_k \times m} \}_{k=1}^d$):
%as the following:
\begin{equation} \label{eq:lff_definition}
	f(\ve x) 
	\;\;:=\;\; 
		\ve a^\top \ve \basefun(\ve x)
	\;\;:=\;\; 
		\ve a^\top \Big[\smallprod{k=1}{d} 
		\ve\basefun^k(x_k)\Big] 
	\;\;:=\;\; 
		\smallsum{i=1}{m} a_i \smallprod{k=1}{d} 
		\smallsum{j=1}{m_k} B^k_{ji} \,
		\indifun_{j}^k(x_k)
	%\;\;\in\;\; \LFF{m} 
	\,.
\end{equation}
LFF are formally defined in \ref{sec:appDefinition}. 
In short,
a basis function $\basefun_i$ is the {\em point-wise product}
of one-dimensional functions $\basefun_i^k$ in each input dimension $k$.
These are themselves constructed as linear functions of a corresponding 
one-dimensional base $\{\indifun_j^k\}_{j=1}^{m_k}$ over that dimension
and ideally can approximate arbitrary functions,
e.g.~Fourier bases or Gaussian kernels.
Although each factored function $\basefun_i$ is very restricted,
a linear combination of them can be very powerful:
\begin{Corollary} \label{th:function_space}
Let $\Set X_k$ be a bounded continuous set and 
$\indifun_j^k$
the $j$'th Fourier base over $\Set X_k$.
In the limit of $m_k \to \infty, 
\forall k \in \{1,\ldots,d\},$
holds $\LFF{\infty} = \Lx$.
\end{Corollary}
Strictly this holds in the limit of infinitely 
many basis functions $\basefun_i$,
but we will show empirically that there exist
close approximations with a small number $m$ of factored functions.
One can make similar statements for other bases 
$\{\indifun_j^k\}_{j=1}^\infty$.
For example, for Gaussian kernels one can show 
that the space $\LFF{\infty}$ is in the limit equivalent to 
the corresponding {\em reproducing kernel Hilbert space} $\hilbert$.

LFF offer some structural advantages over other 
universal function approximation classes 
like neural networks or reproducing kernel Hilbert spaces.
Firstly, the {\em inner product} of two LFF in $\Lx$ can be computed as
products of one-dimensional integrals.
For some bases\footnote{ 
	E.g.~Fourier bases for continuous, and 
	Kronecker-delta bases for discrete variables.
}, these integrals can be calculated analytically without any sampling.
This could in principle break the curse of dimensionality
for algorithms that have to approximate these inner products numerically.
For example, input variables can be 
{\em marginalized} (integrated) out analytically 
(Equation \ref{eq:marginalization} on Page \pageref{eq:marginalization}).
Secondly, the {\em point-wise product} of two LFF is a LFF as well\footnote{
	One can use the trigonometric product-to-sum identities 
	for Fourier bases or the Kronecker delta for discrete bases
	to construct LFF from a point-wise product without changing 
	the underlying basis $\{\{\indifun_i^k\}_{i=1}^{m_k}\}_{k=1}^d$.
} (Equation \ref{eq:pointwise} on Page \pageref{eq:pointwise}). 
See \ref{sec:appDefinition} for details.
These properties are very useful, for example
in {\em belief propagation} \citep{Pearl88} 
and {\em factored reinforcement learning} \citep{Boehmer13b}.

% ------------------------------------------------------------------------------
\subsection{Constrains}
LFF have some degrees of freedom that can impede optimization.
For example, the norm of $\basefun_i \in \FF$
does not influence function $f \in\LFF{m}$, 
as the corresponding linear coefficients 
$a_i$ can be scaled accordingly.
We can therefore introduce the {\em constraints} 
$\|\basefun_i\|_\indi = 1, \forall i$, 
without restriction to the function class.
The factorization of inner products allows us furthermore
to rewrite the constraints as 
$\|\basefun_i\|_\indi = \prod_k 
\|\basefun_i^k\|_{\indi^k} = 1$.
This holds as long as the product is one,
which exposes another unnecessary degree of freedom.
To finally make the solution unique (up to permutation), 
we define the constraints 
as $\|\basefun_i^k\|_{\indi^k} = 1, 
\forall k, \forall i$.
Minimizing some $\cost[f]$ w.r.t.~$f \in \LFF{m}$ 
is thus equivalent to
\vspace{2mm}
\begin{equation} \label{eq:lff_regression}
	\inf_{f \in \LFF{m}} \cost[f] 
	\qquad\quad \text{s.t.} \quad
	\|\basefun_i^k\|_{\indi^k} \;=\; 1 \,, 
	\quad \forall k \in \{1, \ldots, d\}\,, 
	\quad \forall i \in \{1, \ldots, m\}\,.
\end{equation}
The {\em cost function} $\cost[f|\chi,\mu]$ 
of Equation \ref{eq:noisy_regression}
with the constraints in Equation \ref{eq:lff_regression} 
is equivalent to {\em ordinary least squares} (OLS)
w.r.t.~linear parameters $\ve a \in \R^m$.
However, the optimization problem is 
{\em not} convex w.r.t.~the parameter space $\{\mat B^k \in 
\R^{m_k \times m}\}_{k=1}^d$,
due to the nonlinearity of products.

Instead of tackling the global optimization problem
induced by Equation \ref{eq:lff_regression},
we propose a {\em greedy} approximation algorithm\footnote{
	%Similar to {\em projection pursuit} \citep{Friedman74}
	%and {\em matching pursuit} \citep{Mallat93}.
	Similar to {\em projection pursuit} \citep{Friedman74,Mallat93}.
}.
Here we optimize at iteration $\iteration$ one linear basis function 
$\basefun_\iteration =: g =: \prod_k g^k \in \FF$,
with $g^k(x_k) =: \ve b^{k\top} 
\ve \indifun^k(x_k)$, 
at a time, to fit the residual $\mu - f$
between the true {\em mean label} function $\mu \in \Lx$ and 
the current maximum likelihood estimate $f \in \LFF{\iteration-1}$, 
based on all $\iteration-1$ previously constructed factored basis functions
$\{\basefun_i\}_{i=1}^{\iteration-1}$:
\begin{eqnarray}\label{eq:greedy_regression}
\inf_{g \in \FF} \cost[f+g|\chi,\mu]
	&& \qquad \text{s.t.} \quad 
	\|g^k\|_{\indi^k} = 1 \,, 
	\quad \forall k \in \{1,\ldots,d\}\,.
\end{eqnarray}

% ------------------------------------------------------------------------------
\subsection{Regularization} \label{sec:lff_greedy} \label{sec:lff_regularization}
Regression with any powerful function class
requires regularization to avoid over-fitting.
Examples are {\em weight decay} for neural networks \citep{Haykin98}
or parameterized {\em priors} for Gaussian processes.
It is, however, not immediately obvious 
how to regularize the parameters of a LFF
and we will derive a regularization term
from a Taylor approximation of the cost function 
in Equation \ref{eq:noisy_regression}.

We aim to enforce smooth functions, 
especially in those regions our knowledge 
is limited due to a lack of training samples.
This {\em uncertainty} can be expressed as the
{\em Radon-Nikodym derivative}\/\footnote{
	Technically we have to assume that $\indi$ 
	is {\em absolutely continuous} in respect to $\xi$.
	For ``well-behaving'' distributions $\indi$,
	like the uniform or Gaussian distributions
	we discuss in  \ref{sec:appDefinition},
	this is equivalent to the assumption
	that in the limit of infinite samples,
	each sample $\ve z \in \Set X$
	will {\em eventually} be drawn by $\xi$.
} $\rnd: \Set X \to [0, \infty)$
between our factored measure $\indi$ 
(see  \ref{sec:appDefinition})
and the sampling distribution $\xi$.
In the example of a uniform distribution $\indi$,
$\rnd$ corresponds to the reciprocal
of $\xi$'s {\em probability density function},
and therefore reflects our empirical knowledge of the input space.

We use this uncertainty to modulate the 
{\em sample noise distribution} $\chi$ 
in Equation \ref{eq:noisy_regression}.
This means that frequently sampled regions of $\Set X$ 
shall yield low, while scarcely sampled 
regions shall yield high variance.
Formally, we assume $\chi(d\ve x | \ve z)$ to be a 
Gaussian probability measure over $\Set X$ with mean $\ve z$, that is
\begin{equation} \label{eq:noise_def}
	\smallint \chi(d\ve x|\ve z)  
		(\ve x - \ve z) = \ve 0 \,,
	\quad\;\;
	\smallint \chi(d\ve x|\ve z)  
		(\ve x - \ve z)(\ve x -\ve z)^\top 
		= \rnd(\ve z) \cdot \mat\Sigma   \,,
	\quad\;\;
	\forall \ve z \in \Set X \,.
\end{equation}
In the following we assume without loss of generality\footnote{
	Non-diagonal covariance matrices $\mat \Sigma$ can 
	be cast in this framework by projecting the input 
	samples into the eigenspace of $\mat \Sigma$
	(thus diagonalizing the input)
	and use the corresponding eigenvalues $\lambda_k$
	instead of the regularization parameters $\sigma^2_k$'s.
} the {\em covariance matrix} $\mat \Sigma \in \R^{d \times d}$ 
to be diagonal, with the diagonal
elements called $\sigma_k^2 := \Sigma_{kk}$.
Note that $\mat \Sigma$ is modulated by 
the Radon-Nikodym derivative
$\rnd(\ve z), \forall \ve z \in \Set X$.

\begin{Proposition} \label{th:taylor_regression}
Under the assumptions of 
Equation \ref{eq:noise_def}
and a diagonal covariance matrix $\mat \Sigma$,
%with $\sigma^2_k := \Sigma_{kk}, \forall k \in \{1,\ldots,d\}$,
the first order Taylor approximation of
the cost $\cost$ in  
Equation \ref{eq:greedy_regression} is
\begin{equation} \label{eq:lff_approx}
	%\inf_{g \in \FF} \, 
	\acost[g] \quad := \quad
	\underbrace{\|g - (\mu - f)\|_\xi^2 \;}_{\text{\rm sample-noise free cost}}
	+ \;\;\, \smallsum{k=1}{d} \, \sigma_k^2
	\kern-2.2ex\underbrace{\|\smallfrac{\partial}{\partial x_k}g 
		+ \smallfrac{\partial}{\partial x_k}f \|_\indi^2 
	}_{\text{\rm smoothness in dimension~}k} \kern-2.2ex.
	%\;\;\, \text{\rm s.t.} \,
	%\|g^k\|_{\indi^k} = 1 \,, 
	%\, \forall k \in \{1,\ldots,d\} .
\end{equation}
\end{Proposition}
{\bf Proof:} see  \ref{sec:appProofs}
on Page \pageref{proof_taylor_regression}.
\hfill $\Box$

Note that the approximated cost
$\acost[g]$ consists of the sample-noise free cost
(measured w.r.t.~training distribution $\xi$)
and $d$ regularization terms. 
%that each prefer smooth functions in one input dimension.
Each term prefers functions that are smooth in one input dimension
and is measured 
w.r.t.~the factored distribution $\indi$,
e.g.~everywhere equally for a uniform $\indi$.
This enforces smoothness everywhere,
but allows exceptions where enough data is available.
To avoid a cluttered notation, 
we will use the symbol $\subgrad{k}f := 
{\scriptstyle\frac{\partial}{\partial x_k}}f$.

% ++++++++++++++++++++++++++++++++++++++++++++++++++++++++++++++++++++++++++++++
\setcounter{algorithm}{0}
\begin{algorithm*}[t]
\caption{ {\bf (abstract)} 
	\,--\, a detailed version can be found on Page \pageref{alg:regression} }
\begin{algorithmic}
\WHILE { new factored basis function can improve solution }
	\STATE initialize new basis function $g$ as constant function
	\WHILE {optimization improves cost in Equation \ref{eq:lff_approx} }
		\FOR {random input dimension $k$}
			\STATE calculate optimal solution for $g^k$
				without changing $g^l, \forall l \neq k$
		\ENDFOR 
	\ENDWHILE \quad  // new basis function $g$ has converged 
	\STATE add $g$ to set of factored basis functions and solve OLS 
\ENDWHILE \qquad // regression has converged
\end{algorithmic}
\end{algorithm*}

% ++++++++++++++++++++++++++++++++++++++++++++++++++++++++++++++++++++++++++++++

% ------------------------------------------------------------------------------
\subsection{Optimization} \label{sec:lff_greedy}

Another advantage of cost function $\acost[g]$
is that one can optimize one factor function $g^k$
of $g(\ve x) = g^1(x_1) \cdot \ldots \cdot g^d(x_d) \in \FF$ 
at a time, instead of time consuming {\em gradient descend}
over the entire parameter space of $g$.
%$\{\ve b^k \in \R^{m_k}\}_{k=1}^d$.
To be more precise:

\begin{Proposition} \label{th:regression_unique}
If all but one factor $g^k$ are considered constant, 
Equation \ref{eq:lff_approx} has an analytical solution.
If $\{\indifun^k_j\}_{j=1}^{m_k}$ is a Fourier base,
$\sigma_k^2 > 0$ and $\indi \ll \xi$, 
then the solution is also unique.
\end{Proposition}
\vspace{-2mm}
{\bf Proof:} see \ref{sec:appProofs} on Page 
\pageref{proof_regression_unique}. \hfill $\Box$

\vspace{1mm}
\noindent
One can give similar guarantees for other bases,
e.g.~Gaussian kernels.
Note that Proposition \ref{th:regression_unique}
does {\em not} state that the optimization
problem has a unique solution in $\FF$.
Formal convergence statements are not trivial
and empirically the parameters of $g$ do not converge,
but evolve around orbits of equal cost instead.
However, since the optimization of {\em any} $g^k$
cannot increase the cost, any sequence
of improvements will converge to (and stay in) a {\em local minimum}.
This implies a {\em nested} optimization approach,
that is formulated in Algorithm \ref{alg:regression}
on Page \pageref{alg:regression}:
%\vspace{-2mm}
\setlength{\itemsep}{0mm}
\begin{itemize}%[leftmargin=6mm]
\item 
	An {\em inner loop} that optimizes one factored basis function
	$g(\ve x) = g^1(x_1) \cdot\ldots\cdot g^d(x_d)$ 
	by selecting an input dimension $k$ in each iteration and 
	solve Equation \ref{eq:lff_approx} for the corresponding $g^k$.
	A detailed derivation of the optimization steps 
	of the inner loop is given in
	\ref{sec:appInnerLoop} on Page \pageref{sec:appInnerLoop}.
	The choice of $k$ influences 
	the solution in a non-trivial way and 
	further research is needed to build up a rationale
	for any meaningful decision.
	For the purpose of this paper, we assume 
	$k$ to be chosen randomly by permuting the order of updates.
	
	The {\em computational complexity} of the inner loop is
	$\Set O(m_k^2 n + d^2  m_k m)$.
	Memory complexity is $\Set O(d\,m_k m)$, or 
	$\Set O(d\,m_k n)$ with the optional cache
	speedup of Algorithm \ref{alg:regression}.
	The loop is repeated for random $k$ until the cost-improvements 
	of all dimensions $k$ fall below some small $\epsilon$.
\item
	After convergence of the inner loop in (outer) iteration $\iteration$,
	the new basis function is $\basefun_\iteration := g$.
	As the basis has changed, 
	the linear parameters $\ve a \in \R^{\iteration}$
	have to be readjusted by solving 
	the ordinary least squares problem 
	\vspace{-1mm}
	$$
		\ve a \;=\; (\mat \Psi \mat \Psi^\top)^{-1} 
		\mat \Psi \ve y \,,	\;\; \text{with} \;\;
		\Psi_{it} \;:=\; \basefun_i(\ve x_t) \,,
		\; \forall i \in \{1, \ldots, \iteration\} \,,
		\; \forall t \in \{1, \ldots, n\} \,.
	$$
	
	\vspace{-1mm}
	%Convergence of the outer loop is less trivial.
	We propose to stop the approximation when
	the newly found basis function $\basefun_\iteration$ is no longer 
	{\em linearly independent} of the current basis 
	$\{\basefun_i\}_{i=1}^{\iteration-1}$.
	This can for example be tested by comparing the {\em determinant}
	$\det(\frac{1}{n} \mat \Psi \mat \Psi^\top) < \varepsilon$,
	for some very small $\varepsilon$.
\end{itemize}

% ------------------------------------------------------------------------------
\begin{figure}[t]
	%\vspace{-2mm}
	\centering
	\includegraphics[width=0.495\textwidth]{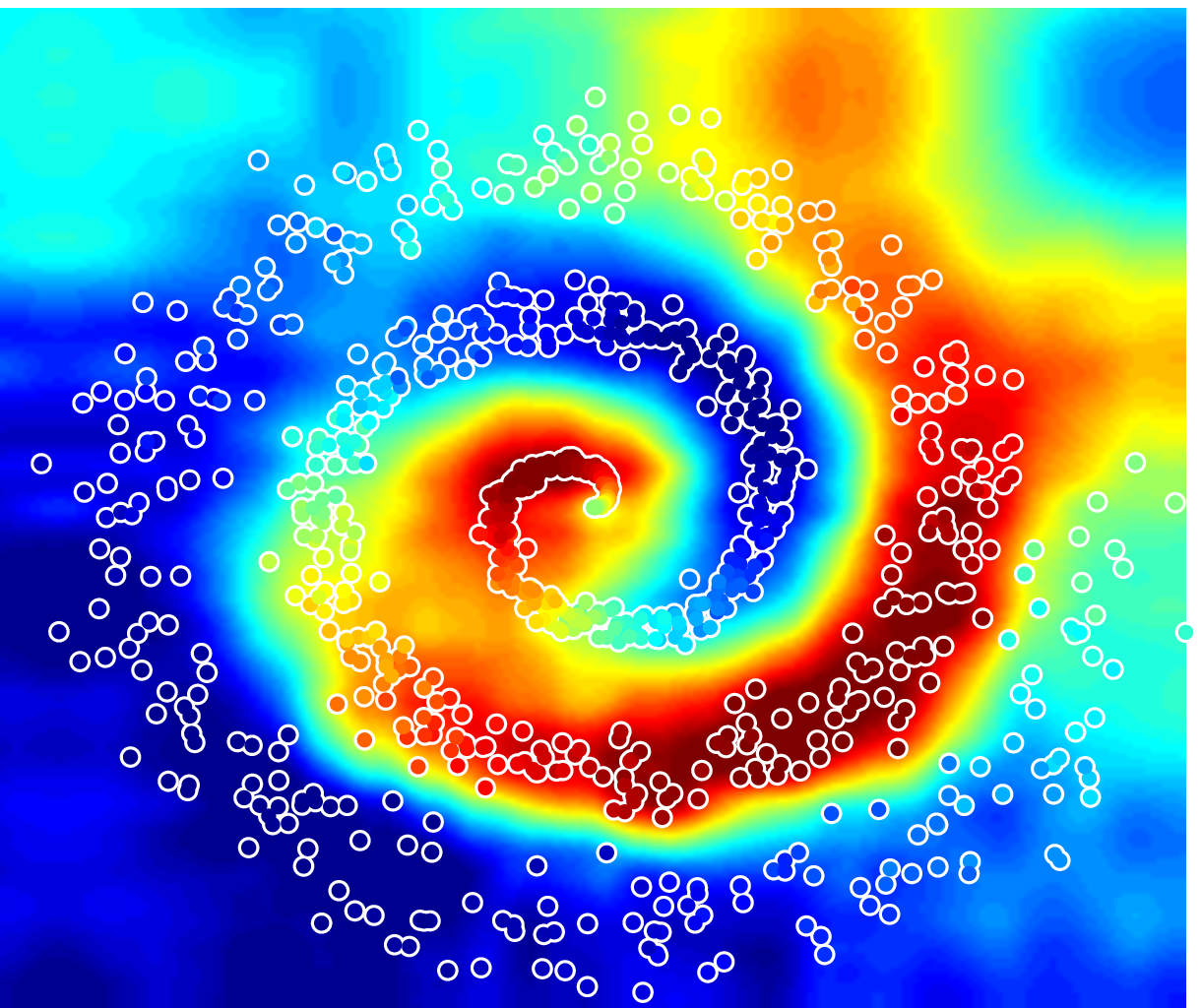}
	\hfill
	\includegraphics[width=0.495\textwidth]{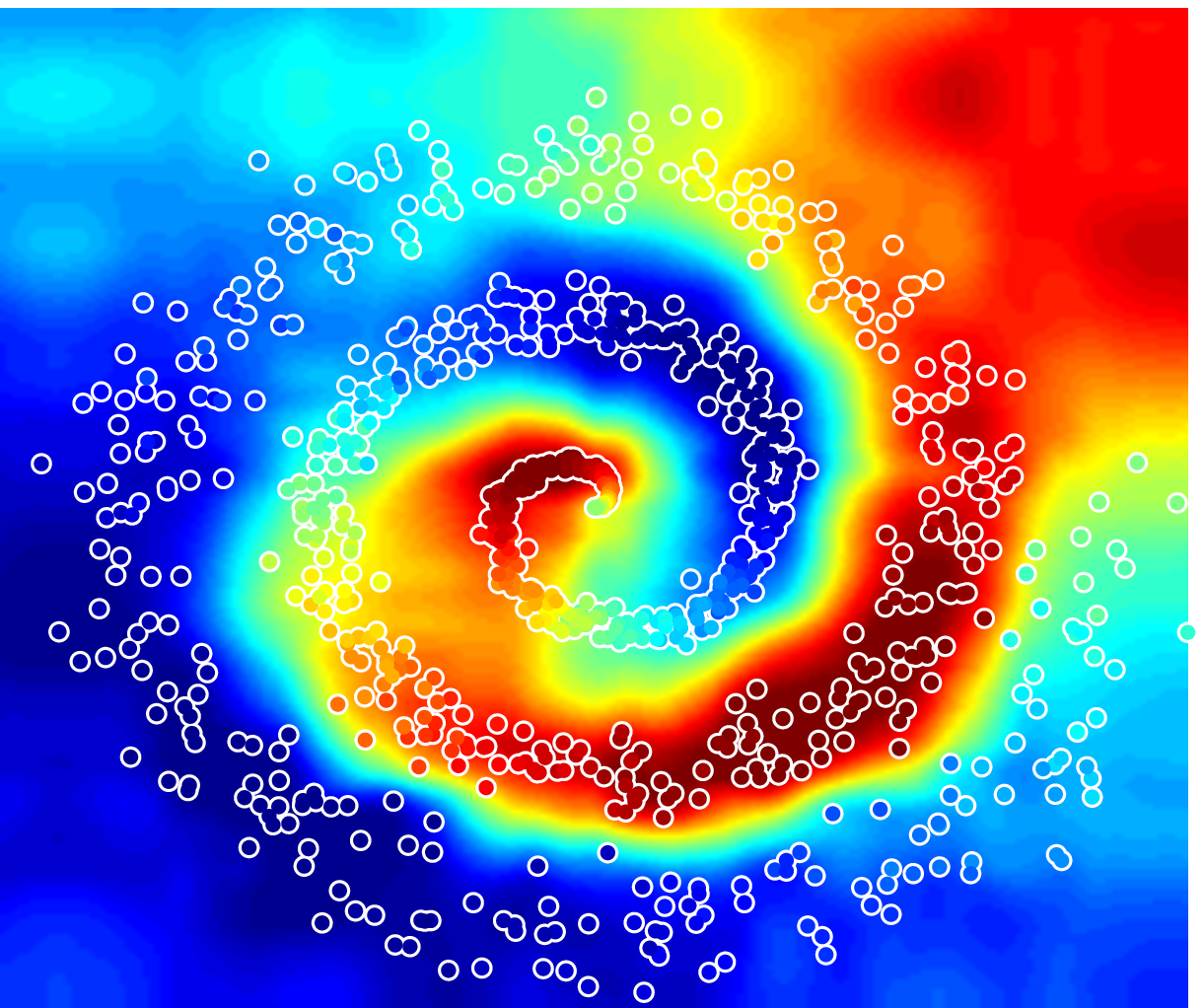}

	\vspace{-1.5mm}
	\caption{ Two LFF functions learned from the same 1000 training samples
			(white circles).
			The color inside a circle represents the training label.
			Outside the circles, the color represents the prediction of 
			the LFF function.
			The differences between both functions
			are rooted in the randomized order
			in which the factor functions $g^k$ are updated.
			Note the similarity of the sampled region, though.}
	\label{fig:fullscreen}
	\vspace{-2mm}
\end{figure}

% ==============================================================================
\section{Empirical evaluation} \label{sec:evaluation}
In this section we will evaluate the novel 
LFF regression Algorithm \ref{alg:regression}.
%on Page \pageref{alg:regression}.
We will analyze its properties on low dimensional toy-data,
and compare its performance with sparse 
and traditional Gaussian processes 
\citep[GP, see][]{Rasmussen06,Bishop06}.

% ------------------------------------------------------------------------------
\subsection{Demonstration} \label{sec:toy}
To showcase the novel Algorithm \ref{alg:regression},
we tested it on an artificial 
two-dimensional regression toy-data set.
The $n = 1000$ training samples were drawn from a noisy spiral
and labeled with a sinus.
The variance of the Gaussian sample-noise grew with the spiral as well:
\begin{equation}
	\ve x_{t} \;=\; 6 \smallfrac{t}{n} \bigg[\begin{array}{cc}
		\cos\big(6 \smallfrac{t}{n} \pi\big) \vspace{1mm}\\
		\sin\big(6 \smallfrac{t}{n} \pi\big)
	\end{array} \bigg]
	+ \Set N\Big(\ve 0, \smallfrac{t^2}{4n^2} \mat I\Big) \,,
	\;\; 
	y_t \;=\; \sin\Big( 4 \smallfrac{t}{n} \pi \Big) \,,
	\;\;
	\forall t \in \{1,\ldots,n\} \,.
\end{equation}
Figure \ref{fig:fullscreen} shows one training set 
plotted over two learned\footnote{
	Here (and in the rest of the paper), 
	each variable was encoded with 50 Fourier 
	cosine bases.
	We tested other sizes as well.
	Few cosine bases result effectively 
	in a low-pass filtered function, whereas every experiment 
	with more than 20 or 30 bases behaved very similar.
	We tested up to $m_k=1000$ bases and did not experience over-fitting.
} functions $f \in \LFF{m}$ with $m=21$ 
and $m=24$ factored basis functions, respectively.
Regularization constants were in both cases 
$\sigma_k^2 = 0.0005, \forall k$.
The differences between the functions 
stem from the randomized order
in which the factor functions $g^k$ are updated.
Note that the sampled regions have similar predictions.
Regions with strong differences, 
for example the upper right corner,
are never seen during training.

In all our experiments, 
Algorithm \ref{alg:regression} always converged.
Runtime was mainly influenced by the input dimensionality 
($\Set O(d^2)$),
the number of training samples ($\Set O(n)$) and 
the eventual number of basis functions ($\Set O(m)$).
The latter was strongly correlated with approximation quality,
i.e., bad approximations converged fast.
Cross-validation was therefore able to find good parameters efficiently
and the resulting LFF were always very similar near the training data.

% ------------------------------------------------------------------------------
\subsection{Evaluation} \label{sec:bench}
We compared the regression performance of LFF and GP 
with cross-validation
on five regression benchmarks from
the {\em UCI Manchine Learning Repository}\/\footnote{
	\url{https://archive.ics.uci.edu/ml/index.html}
}:
\vspace{-1.5mm}
\setlength{\itemsep}{0mm}
\begin{itemize}
\item The {\em concrete compressive strength} data set
	\citep[{\em concrete},][]{Yeh98} consists of $n=1030$ samples with 
	$d=8$ dimensions describing various concrete mixture-components.
	The target variable\, is the real-valued 
	compression strength of the mixture	after it hardened.
\item The {\em combined cycle power plant} data set 
	\citep[{\em ccpp},][]{Tuefekci14} consists of $n=9568$ samples with
	$d=4$ dimensions describing 6 years worth of measurements from
	a combined gas and steam turbine.
	The real-valued target variable is the energy output of the system.
\item The {\em wine quality} data set \citep{Cortez09} consists of
	two subsets with $d=11$ dimensions each, 
	which describe physical attributes
	of various white and red wines: 
	the set contains $n=4898$ samples of {\em white wine}
	and $n=1599$ samples of {\em red wine}.
	The target variable is the estimated wine quality on a 
	discrete scale from 0 to 10.
\item The {\em yacht hydrodynamics} data set 
	\citep[{\em yacht},][]{Gerritsma81}
	consists of $n=308$ samples with $d=6$ dimensions 
	describing parameters of the {\em Delft yacht hull} ship-series.
	The real-valued target variable is the residuary resistance 
	measured in full-scale experiments. 
\end{itemize}
\vspace{-1.75mm}
To demonstrate the advantage of factored basis functions,
we also used the 2d-spiral toy-data set of the previous section
with a varying number of additional input dimensions.
Additional values were drawn i.i.d.~from a Gaussian distribution
and are thus independent of the target labels.
As the input space $\Set X$ grows, kernel methods will
increasingly face the {curse of dimensionality} during training.

Every data-dimension (except the labels) have been
translated and scaled to zero mean and unit-variance before training.
Hyper-parameters were chosen w.r.t.~the mean
of a 10-fold cross-validation.
LFF-regression was tested for the uniform noise-parameters 
$\sigma_k^2 = 10^{u}, \forall k$, 
with $u \in \{-10, -9.75, -9.5, \ldots, 10\}$,
i.e.~for 81 different hyper-parameters.
GP were tested with Gaussian kernels 
$\kernel(\ve x,\ve y) = \exp(-\frac{1}{2\bar\sigma^2} \|\ve x - \ve y\|_2^2)$
using kernel parameters $\bar\sigma = 10^{v}$, 
with $v \in \{-1, -0.75, -0.5, \ldots, 3\}$,
and prior-parameters $\beta = 10^{w}$ 
\citep[see][for the definition]{Bishop06},
with $w \in \{-2, -1, \ldots, 10\}$,
i.e.~for 221 different hyper-parameter combinations.
The number of support vectors in 
standard GP equals the number of training samples.
As this is not feasible for larger data sets,
we used the MP-MAH algorithm \citep{Boehmer12}
to select a uniformly distributed subset of 2000 training samples
for sparse GP \citep{Rasmussen06}.

\begin{figure}[t]
\vspace{-2mm}
\centering
\includegraphics[width=0.495\textwidth]{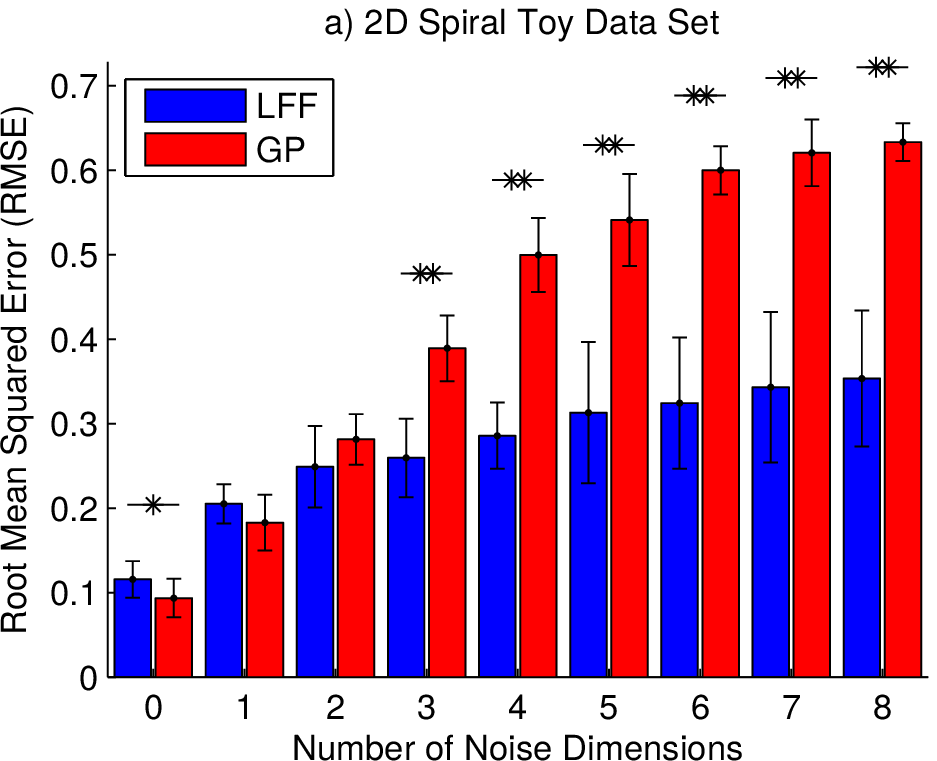}
\hfill
\includegraphics[width=0.48\textwidth]{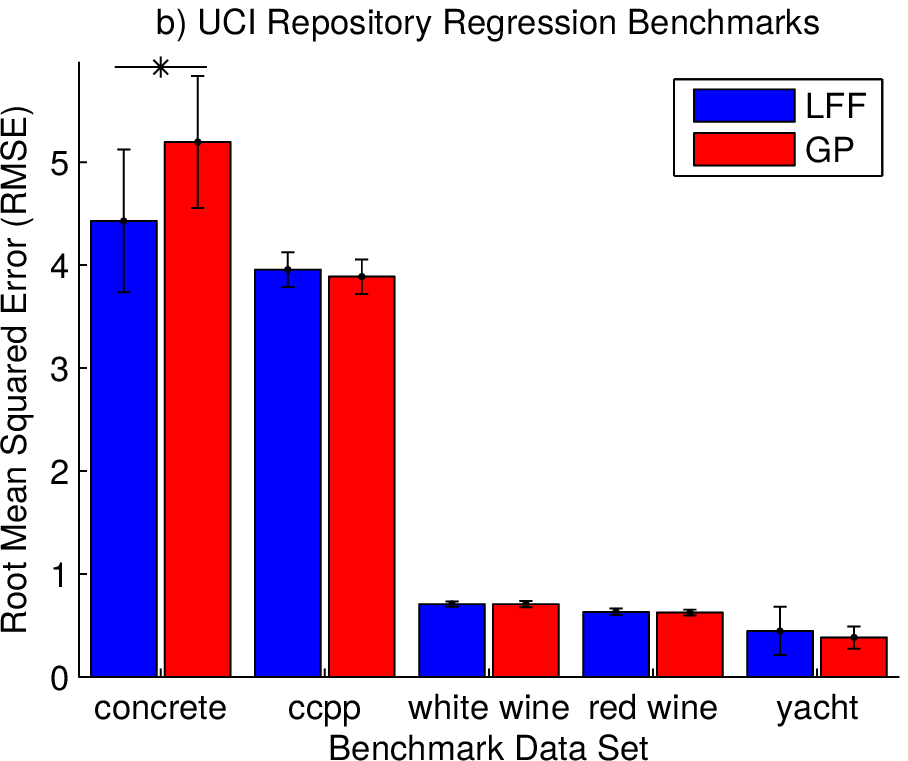}
\vspace{-2mm}
\caption{Mean and standard deviation within a 10-fold 
		cross-validation of a) the toy data set 
		with additional independent noise input dimensions
		and b) all tested UCI benchmark data sets.
		The stars mark {\em significantly} different 
		distribution of RMSE over all folds 
		in both a {\em paired-sample t-test} 
		and a {\em Wilcoxon signed rank test}.
		Significance levels are: one star $p < 0.05$, two stars $p < 0.005$.}
\label{fig:benchmark}
\end{figure}

Figure \ref{fig:benchmark}a demonstrates the advantage of 
factored basis functions over kernel methods during training.
The plot shows the {\em root mean squared errors}\footnote{
	RMSE is not a common performance metric for GP,
	which represent a {\em distribution} of solutions.
	However, RMSE reflect the objective of regression
	and are well suited to compare our algorithm 
	with the {\em mean} of a GP.
} (RMSE) of the two dimensional spiral toy-data set
with an increasing number of independent noise dimensions.
GP solves the initial task better,
but as the size of the input space $\Set X$ grows,
it clearly succumbs to the curse of dimensionality.
LFF, on the other hand, 
significantly overtake GP from 3 noise dimensions on,
as the factored basis functions appear to be
less affected by the curse.
Another difference to GP is that 
decreasing performance automatically 
yields less factored basis functions
(from $19.9 \pm 2.18$ with 0, 
to $6.3 \pm 0.48$ bases with 8 noise dimensions).

\begin{table}[t]
\caption{10-fold cross-validation {\em root mean squared error} (RMSE)
	for benchmark data sets with 
	$d$ dimensions and $n$ samples, 
	resulting in $m$ {\em basis functions}.
	Significantly smaller RMSE are in bold.}
\label{tab:benchmark}
\begin{center}
\renewcommand{\tabcolsep}{1.5mm}
\renewcommand{\arraystretch}{1.2}
\begin{tabular}{|lcc|cc|cc|}
\hline
{\bf DATA SET} & $\ve d$ & $\ve{n}$ & 
{\bf RMSE LFF} & {\bf RMSE GP} & 
$\ve m$ {\bf LFF} & $\ve m$ {\bf GP}
\\ \hline 
%Concrete Compressive Strength
Concrete & 8 & 1030
	& $\mathbf{4.429 \pm 0.69}$ & $5.196 \pm 0.64$ 
	& $4.2 \pm 0.8$ & $927$ 
	\\
%Combined Cycle Power Plant 
CCPP & 4 & 9568
	& $3.957 \pm 0.17$ & $3.888 \pm 0.17$
	& $8.8 \pm 2.0$ & 2000 
	\\
%Wine Quality (white) 
White Wine & 11 & 4898
	& ${0.707} \pm 0.02$ & $0.708 \pm 0.03$
	& $4.2 \pm 0.4$ & 2000
	 \\
%Wine Quality (red) 
Red Wine & 11 & 1599
	& $0.632 \pm 0.03$ & ${0.625} \pm 0.03$ 
	& $4.7 \pm 0.7$ & 1440 
	\\
%Yacht Hydrodynamics 
Yacht & 6 & 308
	& $0.446 \pm 0.23$ & ${0.383} \pm 0.11$
	& $4.2 \pm 0.6$ & 278 
	\\
\hline
\end{tabular}
\end{center}
\end{table}

Figure \ref{fig:benchmark}b and Table \ref{tab:benchmark}
show that our LFF algorithm performs on all evaluated
real-world benchmark data sets comparable to (sparse) GP.
RMSE distributions over all folds 
were statistically indistinguishable,
except for an advantage of LFF regression in
the concrete compressive strength data set
($p < 0.01$ in a {\em t-test} 
and $p < 0.02$ in a {\em signed rank test}).
As each basis function requries many iterations to converge,
LFF regression 
runs considerably longer than standard approaches.
However, LFF require between 3 and 12 factored basis functions
to achieve the {\em same} performance as GP with 278-2000 
kernel basis functions.

%\newpage
% ==============================================================================
\section{Discussion} \label{sec:discussion}
% Representations/summary
We presented a novel algorithm for regression,
which constructs factored basis functions during training.
As {\em linear factored functions} (LFF) can
in principle approximate any function in $\Lx$,
a regularization is necessary to avoid over-fitting.
Here we rely on a regularization scheme that 
has been derived from a Taylor approximation 
of the least-squares cost function with virtual sample-noise.
RMSE performance appears comparable to Gaussian processes
on real-world benchmark data sets,
but the factored representation is considerably more compact
and seems to be less affected by distractors.

At the moment, LFF optimization faces two challenges.
(i) The optimized cost function is not convex, 
but the local minimum of the solution may be controlled  
by selecting the next factor function to optimize.
This will require further research,
but may also allow some performance guarantees.
(ii) The large number of inner-loop iterations 
make the algorithm slow.
This problem should be solved by addressing (i), but 
finding a trade-off between approximation quality 
and runtime may also provide a less compact shot-cut with 
similar performance.

Preliminary experiments also demonstrated the viability
of LFF in a {\em sparse regression} approach.
Sparsity refers here to a limited number of 
input-dimensions that affect the prediction,
which can be implemented by adjusting the
sample-noise parameters $\sigma^2_k$ 
during training for each variable $\Set X_k$ individually.
This is of particular interest, as 
factored functions are ideally suited to represent sparse functions
and are in principle {\em unaffected} 
by the curse of dimensionality in function representation.
Our approach modified the cost function to
enforce LFF functions that were constant 
in all noise-dimensions.
We did not include our results in this paper, 
as choosing the first updated factor functions $g^k$ poorly
resulted in basis functions that 
rather fitted noise than predicted labels.
When we enforce sparseness,
this initial mistake can afterwards no longer
be rectified by other basis functions,
in difference to the presented Algorithm \ref{alg:regression}.
However, if this can be controlled
by a sensible order in the updates,
the resulting algorithm should be much faster
and more robust than the presented version.

There are many application areas that may exploit
the structural advantages of LLF.
In {\em reinforcement learning} \citep{Kaelbling96}, 
one can exploit the factorizing inner products
to break the curse of dimensionality of the state space \citep{Boehmer13b}.
Factored transition models also need to be learned from experience, 
which is essentially a sparse regression task.
Another possible field of application are
{\em junction trees} \citep[for Bayesian inference, see e.g.~][]{Bishop06}
over continuous variables,
where sparse regression may estimate the conditional probabilities.
In each node one must also marginalize out variables,
or calculate the point-wise product over multiple functions.
Both operations can be performed analytically with LFF,
the latter at the expense of more 
basis functions in the resulting LFF.
However, one can use our framework to 
{\em compress} these functions after multiplication.
This would allow junction-tree inference 
over mixed continuous and discrete variables.

In summary,
we believe our approach to approximate functions
by constructing non-linear factored basis functions (LFF)
to be very promising.
The presented algorithm performs already comparable 
with Gaussian processes, 
but appears less sensitive to
large input spaces than kernel methods.
We also discussed some potential extensions for sparse regression
that should improve upon that, in particular on runtime,
and gave some fields of application
that would benefit greatly from the
algebraic structure of LFF.

% ==============================================================================
\subsubsection*{Acknowledgments}
The authors thank Yun Shen for his helpful comments.
This work was funded by the {\em German science foundation} (DFG) 
within SPP 1527 {\em autonomous learning}.

% ==============================================================================
\appendix
%\section*{Appendices}
\renewcommand{\thesection}{Appendix \Alph{section}}
% ==============================================================================
\section{LFF definition and properties} \label{sec:appDefinition}
Let $\Set X_k$ denote the subset of $\R$ associated 
with the $k$'th variable of input space $\Set X \subset \R^d$,
such that $\Set X := \Set X_1 \times \ldots \times \Set X_d$.
Let furthermore $\indi$ be a 
{\em factored probability measure} on $\Set X$,
i.e.~$\indi(d\ve x) = \prod_{k=1}^d \indi^k(dx_k),
\int \indi^k(dx_k) = 1, \forall k$.
For example, $\indi^k$ could be uniform or Gaussian
distributions over $\Set X_k$ 
and the resulting $\indi$ would be 
a uniform or Gaussian distribution
over the input space $\Set X$.

A function $g: \Set X \to \R$ is called a {\em factored function}
if it can be written as a product of one-dimensional 
{\em factor functions} $g^k: \Set X_k \to \R$,
i.e.~$g(\ve x) = \prod_{k=1}^{d} g^k(x_k)$.
We only consider factored functions $g$ that are 
twice integrable w.r.t.~measure $\indi$, i.e.~$g \in \Lx$.
Note that not all functions $f \in \Lx$ are factored, though.
Due to {\em Fubini's theorem} 
the $d$-dimensional inner product between 
two factored functions $g,g'\in\Lx$
can be written as the product of $d$ one-dimensional inner products:

\begin{equation*} \label{eq:fubini}
	\wsp{g}{g'}{\indi} 
	=	\int \indi(d\ve x) \, g(\ve x) \, g'(\ve x)
	= \int \smallprod{k=1}{d} \indi^k(dx_k) \, 
		g^k(dx_k) \, g'^k(dx_k)
	= \smallprod{k=1}{d}
		\wsp{g^k}{g'^k}{\indi^k} \,.
\end{equation*}
This trick can be used to solve the integrals 
at the heart of many least-squares algorithms.
Our aim is to {\em learn} factored basis functions $\basefun_i$. 
To this end, 
let $\{\indifun^k_j : \Set X_k \to \R\}_{j=1}^{m_k}$
be a well-chosen\footnote{
	Examples for continuous variables $\Set X_k$
	are Fourier cosine bases 
	$\indifun_j^k(x_k) \sim \cos\big((j-1)\,\pi\,x_k\big)$,
	and Gaussian bases $\indifun_j^k(x_k) = 
	\exp\big(\frac{1}{2\sigma^2}(x_k - s_{k j})^2\big)$.
	Discrete variables may be represented with
	Kronecker-delta bases
	$\indifun_j^k(x_k = i) = \delta_{ij}$.
} (i.e.~universal) basis on $\Set X_k$,
with the space of linear combinations denoted by
$\LFa := \{\ve b^\top \ve\indifun^k | \ve b \in \R^{m_k}\}$.
One can thus approximate factor functions of $\basefun_i$ in $\LFa$, 
i.e., as linear functions
\begin{equation}
	\basefun_i^k(x_k) 
	\quad:=\quad \smallsum{j=1}{m_k} B^k_{ji} \, 
	\indifun^k_j(x_k) 
	\quad\in\quad \LFa \,,
	\qquad \qquad
	\mat B^k \;\; \in \;\; \R^{m_k \times m} \,.
\end{equation}
Let $\FF$ be the space of all factored basis functions $\basefun_i$
defined by the factor functions $\basefun_i^k$ above, 
and $\LFF{m}$ be the space of all 
linear combinations of those $m$ factored basis functions
(Equation \ref{eq:lff_definition}).

{\em Marginalization} of LFF can be performed analytically
with Fourier bases $\indifun^k_j$ and uniform distribution $\indi$
(many other bases can be analytically solved as well): 
\begin{equation} \label{eq:marginalization}
\hspace{-0.75mm}
	\int \indi^l(dx_l) \, f(\ve x) 	=
	\sum_{i=1}^{m} \Big( a_i  
		\smallsum{j=1}{m_l} B^l_{ji} \, 
		\underbrace{\wsp{\indifun_j^l}{1}{\indi^l}
		}_{\text{mean of}~\indifun_j^l}
	\Big) \Big[ \smallprod{k \neq l}{d} \basefun_i^k\Big]
	\stackrel{\text{Fourier}}{=}
	\sum_{i=1}^{m}  \underbrace{ a_i B^l_{1i}
	}_{\text{new}~a_i} 
	\Big[ \smallprod{k \neq l}{d} \basefun_i^k\Big] \,. \hspace{-5mm}
\end{equation}
\vspace{-3mm}

\noindent
Using the trigonometric {\em product-to-sum} identity
$\cos(x) \cdot \cos(y) = \frac{1}{2} \big(\cos(x-y) + \cos(x+y) \big)$,
one can also compute the point-wise product 
between two LFF $f$ and $\bar f$ with cosine-Fourier base
(solutions to other Fourier bases are less elegant):
\begin{equation} \label{eq:pointwise} \hspace{-1.5mm}
\begin{array}{rcl}
	\tilde f(\ve x) &:=&
	f(\ve x) \cdot \bar f(\ve x) \vspace{-4.5mm}\\ 
	 &\stackrel{\text{Fourier}}{=}&
	{\displaystyle\sum\limits_{i,j=1}^{m \bar m}}
	\underbrace{a_i \, \bar a_j}_{\text{new}~\tilde a_{t}} 
	{\displaystyle \prod\limits_{k=1}^d \sum\limits_{l=1}^{2m_k}}
	\Big( \overbrace{ 
		\smallfrac{1}{2} \smallsum{q=1}{l-1}
			B^k_{qi} \, \bar B^k_{(l-q)j} 
		+ \smallfrac{1}{2} \smallsum{q=l+1}{m_k}
			B^k_{qi} \, \bar B^k_{(q-l)j} 
	}^{\text{new}~\tilde B^k_{lt}} \Big) \, \indifun^k_l(x_k) \,, \hspace{-5mm}
\end{array}
\end{equation} \vspace{-3mm}

\noindent
where $t := (i-1) \, \bar m + j$, 
and $B^k_{ji} := 0, \forall j > m_k$, 
for both $f$ and $\bar f$.
Note that this increases the number of basis functions 
$\tilde m = m \bar m$, and the number of bases $\tilde m_k = 2 m_k$
for each respective input dimension.
The latter can be counteracted by {\em low-pass filtering},
i.e., by setting $\tilde B^k_{ji} := 0, \forall j > m_k$.

% ------------------------------------------------------------------------------
\section{Inner loop derivation} \label{sec:appInnerLoop}
Here we will optimize the problem 
in Equation \ref{eq:lff_approx}
for one variable $\Set X_k$ at a time,
by describing the update step $g^k \leftarrow g'^k$.
This is repeated with randomly chosen variables $k$,
until convergence of the cost $\acost[g]$,
that is, until all possible updates 
decrease the cost less than some small $\epsilon$.

Let in the following
$\mat C^k := \wsp{\ve\indifun^k}
{\ve\indifun^{k\top}}{\indi^k}$ 
and $\mat{\dot C}^k := \wsp{\subgrad{k}\ve\indifun^k}
{\subgrad{k}\ve\indifun^{k\top}}{\indi^k}$
denote covariance matrices,
and $\ve R^k_l := 
\ddb{k} \wsp{\subgrad{l}g}{\subgrad{l}f}{\indi}$
denote the derivative of one regularization term.
Note that for some choices of bases 
$\{\indifun^k_j\}_{j=1}^{m_k}$,
one can compute the covariance matrices 
analytically before the main algorithm starts,
e.g.~Fourier cosine bases have $C^k_{ij} = \delta_{ij}$ 
and $\dot C^k_{ij} = (i-1)^2 \, \pi^2 \, \delta_{ij}$.

\noindent
The approximated cost function in Equation \ref{eq:lff_approx} is
$$
\acost[g] \;\; = \;\;
	\|g\|_\xi^2 - 2 \wsp{g}{\mu-f}{\xi} + \|\mu - f\|_\xi^2 
	+ \smallsum{k=1}{d} \sigma_k^2
	\Big( \|\subgrad{k}g\|_\indi^2 
	+ 2 \wsp{\subgrad{k}g}{\subgrad{k}f}{\indi}
	+ \|\subgrad{k}f\|_\indi^2  \Big) \,.
$$
The non-zero gradients of all inner products of this equation 
w.r.t.~parameter vector $\ve b^k \in \R^{m_k}$ are
\begin{eqnarray*}
\ddb{k} \wsp{g}{g}{\xi} &=& 
	2 \, \wsp{\ve \indifun^k \cdot 
	\smallprod{l\neq k}{} g^l}
	{\smallprod{l\neq k}{} g^l 
	\cdot \ve\indifun^{k\top}}{\xi} \ve b^k  \,, \\
\ddb{k} \wsp{g}{\mu-f}{\xi} &=&
	\wsp{\ve \indifun^k \cdot 
	\smallprod{l\neq k}{} g^l}{\mu - f}{\xi}  \,, \\
\ddb{k} \wsp{\subgrad{l}g}{\subgrad{l}g}{\indi} &=& 
	\ddb{k} \wsp{\subgrad{l}g^l}
	{\subgrad{l}g^l}{\indi^l} 
	\smallprod{s\neq l}{} 
	\overbrace{\wsp{g^s}{g^s}{\indi^s}}^{1}
	\quad=\quad 2 \, \delta_{kl} \, 
	\mat{\dot C}^k \ve b^k \,, \\
\ve R^k_{l} \quad:= \quad
\ddb{k} \wsp{\subgrad{l}g}{\subgrad{l}f}{\indi} &=&
	\left\{\begin{array}{ll}
		\mat{\dot C}^k \mat B^k 
			\Big[\ve a \cdot \smallprod{s\neq k}{}
			\mat B^{s\top} \mat C^s \ve b^s \Big]
			&,\, \text{if}\quad k = l \\
		\mat{C}^k \mat B^k
			\Big[\ve a \cdot 
			\mat B^{l\top} \mat{\dot C}^l \ve b^l \cdot
			\smallprod{s\neq k\neq l}{}
			\mat B^{s\top} \mat C^s \ve b^s \Big]
			&,\, \text{if}\quad k \neq l
	\end{array}\right. \,.
\end{eqnarray*}
Setting this to zero yields 
the unconstrained solution $g_{uc}^k$,
\begin{equation} \label{eq:analytical_b}
	{\ve b}_{uc}^k = \Big(\overbrace{
	\wsp{\ve \indifun^k \cdot \smallprod{l\neq k}{} g^l}
		{\smallprod{l\neq k}{} g^l \cdot \ve\indifun^{k\top}}{\xi}
	+ \sigma^2_k \mat{\dot C}^k }^{
		\text{regularized covariance matrix} \;\; \mat{\bar C}^k}
	\Big)^{\hspace{-1mm}-1} \hspace{-0.5mm}
	\Big( \wsp{\ve \indifun^k \cdot 
	\smallprod{l\neq k}{} g^l}{\mu - f}{\xi}
	- \smallsum{l=1}{d} \ve R^k_{l} \, \sigma_l^2 
	\Big) \,.
\end{equation}
However, these parameters do not satisfy to the constraint
$\|g'^k\|_{\indi^k} \shallbe 1$, 
and have to be normalized:
\begin{equation} \label{eq:normalization}
	\ve b'^k \quad:=\quad 
	\frac{{\ve b}_{uc}^k}{\|g_{uc}^k\|_{\indi^k}}
	\quad=\quad \frac{{\ve b}_{uc}^k}
	{\sqrt{{\ve b}_{uc}^{k\top} \mat C^k {\ve b}_{uc}^k}}  \,.
\end{equation}
The inner loop finishes when for all $k$ the improvement\footnote{
	Anything simpler does not converge, 
	as the parameter vectors often 
	evolve along chaotic orbits in $\R^{m_k}$.
} from $g^k$ to $g'^k$ drops below some 
very small threshold $\epsilon$,
i.e.~$\acost[g] - \acost[g'] < \epsilon$.
Using $g'^l = g^l, \forall l \neq k$,
one can calculate the left hand side:
\begin{eqnarray}
\acost[g] - \acost[g']
&=&
	\|g\|^2_\xi - \|g'\|_\xi^2
	- 2 \wsp{g-g'}{\mu-f}{\xi} \nonumber \\
&& + \smallsum{l=1}{d} \sigma_l^2 \Big[
	\underbrace{\|\subgrad{l}g\|_\indi^2}_{
		\ve b^{l\top} \mat{\dot C}^l \ve b^l}
	- \underbrace{\|\subgrad{l}g'\|_\indi^2}_{
		\ve b'^{l\top} \mat{\dot C}^l \ve b'^l}
	- 2 \underbrace{\wsp{\subgrad{l}g-\subgrad{l}g'}
		{\subgrad{l}f}{\indi}}_{
		(\ve b^k - \ve b'^k)^\top \ve R^k_l
	} \Big] \label{eq:inner_loop_converge} \hspace{2.5cm}
\end{eqnarray}
$$	
\hspace{1cm}
= 2 \wsp{g-g'}{\mu-f}{\xi}
	+ \ve b^{k\top} \mat{\bar C}^k \ve b^{k\top}
	- \ve b'^{k\top} \mat{\bar C}^k \ve b'^{k\top}
	- 2 (\ve b^k - \ve b'^k)^\top \Big(
		\smallsum{l=1}{d} \ve R^k_l \sigma_l^2 \Big) \,.
	\nonumber
$$

% ------------------------------------------------------------------------------
\section{Proofs of the propositions} \label{sec:appProofs}

% ------------------------------------------------------------------------------
\paragraph{Proof of Proposition \ref{th:taylor_regression}:}
\label{proof_taylor_regression}
The 1st Taylor approximation 
of any $g, f \in \Lt{\Set X, \xi\chi}$ around $\ve z \in \Set X$ 
is $f(\ve x) = f(\ve z + \ve x - \ve z) 
\approx f(\ve z) + (\ve x - \ve z)^\top \ve \grad f(\ve z)$.
For the Hilbert space $\Lt{\Set X, \xi\chi}$ 
we can thus approximate:
\begin{eqnarray*}
	\wsp{g}{f}{\xi\chi} 
	&=&
	\int \xi(d\ve z) \int \chi(d\ve x|\ve z) \, g(\ve x) \, f(\ve x) \\
	&\approx& \int \xi(d\ve z) \Big( g(\ve z) \, f(\ve z) 
		\smallint \overbrace{\xi(d\ve x|\ve z) }^{1}
	+ g(\ve z) \, \smallint\overbrace{\chi(d\ve x|\ve z) \, 
		(\ve x - \ve z)}^{\ve 0 \text{ due to 
			(eq.\ref{eq:noise_def})}}\/^\top \ve \grad f(\ve z)  \\
	&& + \smallint\underbrace{\chi(d\ve x|\ve z) \, 
		(\ve x - \ve z)}_{\ve 0 \text{ due to 
			(eq.\ref{eq:noise_def})}}\/^\top \ve \grad g(\ve z) \, f(\ve z)
	+ \ve\grad g(\ve z)^\top  
			\smallint \underbrace{\chi(d\ve x| \ve z) \, 
			(\ve x - \ve z)(\ve x - \ve z)^\top}_{
			{\scriptscriptstyle\frac{\indi}{\xi}}(\ve z) \cdot \mat\Sigma 
				\text{ due to (eq.\ref{eq:noise_def})}}
		\ve\grad f(\ve z) \Big)  \\
	&=& \wsp{g}{f}{\xi}	+ \smallsum{k=1}{d} \sigma^2_k \,
		\wsp{\subgrad{k}g}{\subgrad{k}f}{\indi} \,. 
\end{eqnarray*}
Using this twice and the zero mean assumption 
(Eq.~\ref{eq:noise_def}), we can derive:
\begin{eqnarray*}
	\inf_{g\in\FF} \cost[f+g|\chi,\mu]  
	&\equiv& \inf_{g\in\FF} \;
		\iint \xi(d\ve z) \, \chi(d\ve x|\ve z) 
		\Big(g^2(\ve x) - 2 \, g(\ve x) \, 
		\big(\mu(\ve z) - f(\ve x)\big)\Big) \nonumber\\
	& = & \inf_{g\in\FF} \;
		\wsp{g}{g}{\xi\chi} + 2 \wsp{g}{f}{\xi\chi}
		- 2 \int \xi(d\ve z) \, \mu(\ve z) 
		\int \chi(d\ve x| \ve z) \, g(\ve x)  \\
	& \approx & \inf_{g\in\FF} \;
		\wsp{g}{g}{\xi} - 2 \wsp{g}{\mu-f}{\xi} 
		+ \smallsum{k=1}{d} \sigma^2_k
		\Big( \wsp{\subgrad{k} g}{\subgrad{k}g}{\indi} 
			+ 2\wsp{\subgrad{k}g}{\subgrad{k}f}{\indi} \Big) \\
	& \equiv &  \inf_{g\in\FF} \;
		\|g - (\mu - f)\|_\xi^2 \; + 
		\smallsum{k=1}{d} \sigma_k^2
		\|\subgrad{k}g + \subgrad{k}f \|_\indi^2 
		\quad = \quad \acost[g] \,.
\end{eqnarray*}

\vspace{-4mm}
\hfill $\Box$

% ------------------------------------------------------------------------------
\paragraph{Proof of Proposition \ref{th:regression_unique}:}
\label{proof_regression_unique}
The analytical solution to the optimization problem 
in Equation \ref{eq:lff_approx}
is derived in \ref{sec:appInnerLoop}
and has a unique solution if the matrix $\mat{\bar C}^k$,
defined in Equation \ref{eq:analytical_b}, is of full rank:
$$
\mat{\bar C}^k \qquad:=\qquad
	\wsp{\ve \indifun^k \cdot 
		\smallprod{l\neq k}{} g^l}
	{\smallprod{l\neq k}{} g^l \cdot
		\ve\indifun^{k\top}}
	{\xi} \quad+\quad 
	\sigma^2_k \mat{\dot C}^k
$$
For Fourier bases the matrix $\mat{\dot C}^k$ is diagonal,
with $\dot C^k_{11}$ being the only zero entry.
$\mat{\bar C}^k$ is therfore full rank if $\sigma_k^2 > 0$
and $\bar C^k_{11} > 0$.
Because $\indi$ is {\em absolutely continuous} in respect to $\xi$,
the constraint $\|g^l\|_\indi=1, \forall l,$ implies that
ther exist no $g^l$ that is zero on {\em all} training samples.
\qquad\quad
As the first Fourier base is a constant,
$\wsp{\indifun_1^k \cdot \prod_{l \neq k} g^l}
{\prod_{l \neq k} g^l \cdot \indifun_1^k}{\xi} > 0$
and the matrix $\mat{\bar C}^k$ is therefore of full rank.
\hfill $\Box$

% ==============================================================================
{\small
\bibliographystyle{plainnat}
\bibliography{bibliography}
}

% ++++++++++++++++++++++++++++++++++++++++++++++++++++++++++++++++++++++++++++++
\setcounter{algorithm}{0}
\begin{algorithm}[h]
\caption{{\bf (detailed)} \; -- \; LFF-Regression}
\label{alg:regression}
\begin{algorithmic}
\STATE {\bf Input:} $\quad \mat X \in \R^{d \times n}, 
		\quad \ve y \in \R^n \,,
		\quad \ve \sigma^2 \in \R^d\,
		\quad \epsilon, \varepsilon \in \R$
\vspace{1mm}
\STATE $\mat C^k := \wsp{\ve\indifun^k}
		{\ve\indifun^k}{\indi^k} \,, \quad
		\mat{\dot C}^k := \wsp{\grad{\ve\indifun}^k}
		{\grad\ve\indifun^k}{\indi^k} \,, \quad
		\forall k$ \hfill // analytically computed covariances
\vspace{1mm}
\STATE $\Phi^k_{jt} := \ve\indifun_j^k(X_{k t}) \,,
		\quad \forall k \,, \; \forall j \,, \; \forall t$
		\hfill // optional cache of sample-expansion
\vspace{1mm}
\STATE $\ve f := \ve 0 \in \R^n; 
		\quad\ve a := \emptyset \,; 
		\quad \mat B^k := \emptyset \,, \;\; \forall k \,;
		\quad \mat\Psi := \infty$
		\hfill // initialization of empty $f \in \LFF{0}$
\WHILE { $\det\Big(\frac{1}{n} \mat\Psi \mat\Psi^\top\Big) > \varepsilon$ }
	\STATE $\ve b^k := 
			\ve 1^k \in \R^{m_k} \,, \;\; \forall k \,;
			\quad \ve g^k := \ve 1 \in \R^n \,, \;\; \forall k $
			\hfill// initialize all $g^k$
			with constant functions
	\STATE $\ve h := \ve \infty \in \R^d$
			\hfill // initialize estimated improvement
	\vspace{1mm}
	\WHILE {$\max(\ve h) > \epsilon$}
		\vspace{1mm}
		\FOR {$k$ {\bf in} randperm($1,\ldots,d$)}
			\vspace{1mm}
			\STATE $\ve R_k := \mat{\dot C}^k \mat B^k \,
					[\ve a \cdot \smallprod{s \neq k}{}
					\mat B^{s\top} \mat C^s \ve b^s ]$
					\hfill // $\ve R_k = \frac{\partial}
					{\partial \ve b^k} \wsp{\subgrad{k}g}
					{\subgrad{k}f}{\indi}$
			\STATE $\ve R_l := \mat{C}^k \mat B^k \,
					[\ve a \cdot 
					\mat B^{l\top} \mat{\dot C}^l \ve b^l \cdot
					\kern-1.5ex\smallprod{s \neq k 
					\neq l}{}\kern-1ex
					\mat B^{s\top} \mat C^s \ve b^s ] \,,
					\quad \forall l \neq k$
					\hfill // $\ve R_l = \frac{\partial}
					{\partial \ve b^k} \wsp{\subgrad{l}g}
					{\subgrad{l}f}{\indi}$
			\STATE $\mat{\bar C} \; := 
					\mat \Phi^k \Big[ \mat \Phi^{k\top} \cdot
					\smallprod{l\neq k}{} (\ve g^l)^2 \, \ve 1^\top
					\Big] \;+\; \sigma_k^2 \, \mat{\dot C}^k$
					\hfill // approx.~regularized cov.~matrix
					(eq.~\ref{eq:analytical_b})
			\STATE $\ve b' := \mat{\bar C}^{-1} 
					\Big( \mat \Phi^k \Big[
					(\ve y - \ve f) \cdot \smallprod{l\neq k}{}
					\ve g^l	\Big] \;-\; \mat R \ve \sigma^2 \Big)$
					\hfill // $g'$ approx.~residual $\mu - f$
					(eq.~\ref{eq:analytical_b})
			\STATE $\ve b' := \ve b'  \; / \;
					\sqrt{\ve b'^{\top} \mat C^k \ve b'}$
					\hfill // normalize $g' \in \LFa$ 
					(eq.~\ref{eq:normalization})
			\vspace{1mm}
			\STATE $h_k := \smallfrac{2}{n} \, 
					(\ve b^k - \ve b')^\top
					\Big( \mat \Phi^k \Big[(\ve y - \ve f) \cdot
					\smallprod{l\neq k}{} \ve g^l \Big] \Big)$
					\hfill // approximate $2 \wsp{g-g'}{\mu-f}{\xi}$
			\STATE $h_k := h_k + 
					\ve b^k \bar{\mat C} \ve b^k
					- \ve b' \bar{\mat C} \ve b'
					- 2 (\ve b^k - \ve b')^\top 
					  \mat R \ve \sigma^2$
					\hfill // calculate improvement 
					(eq.~\ref{eq:inner_loop_converge})
			\vspace{2mm}
			\STATE $\ve b^k := \ve b' \,;
					\quad \ve g^k := \mat \Phi^{k\top} \ve b^k$
					\hfill // update factor function $g^k$
			\vspace{1mm}
		\ENDFOR \;\,\quad // end function $g^k$ update
		\vspace{1mm}
	\ENDWHILE  \,\quad // end inner loop: cost function 
				converged and thus $g$ optimized
	\STATE ${\mat B}^k := [{\mat B}^k, \ve b^k] 
			\,, \;\; \forall k \,;
			\quad \mat \Psi := \Big[ \smallprod{k=1}{d} 
			\mat B^{k\top} \mat \Phi^k \Big]$
			\hfill // adding $g$ to the bases functions of $f$
	\STATE $\ve a := \big(\mat\Psi \mat\Psi^\top)^{-1} \mat \Psi \ve y \,;
			\qquad \ve f := \mat \Psi^\top \ve a$
			\hfill // project $\mu$ onto new bases
	\vspace{1mm}			
\ENDWHILE \qquad // end outer loop: new base $g$ no longer
			linear independent and thus $f \approx \mu$
\vspace{1mm}
\STATE {\bf Output:} $\quad \ve a \in \R^m, \quad 
	\{ \mat B^k \in \R^{m_k \times m} \}_{k=1}^d$
	\hfill // return parameters of $f \in \LFF{m}$
\end{algorithmic}
\end{algorithm}
% ++++++++++++++++++++++++++++++++++++++++++++++++++++++++++++++++++++++++++++++

% ==============================================================================
\end{document}